\documentclass[conference]{IEEEtran}
\IEEEoverridecommandlockouts
\usepackage{cite}
\usepackage{amsmath,amssymb,amsfonts}
\usepackage{algorithmic}
\usepackage{graphicx}
\usepackage{textcomp}
\usepackage{xcolor}

\usepackage{comment}
\usepackage[english]{babel}
\usepackage[colorlinks=true, linkcolor=blue, citecolor=blue, urlcolor=blue]{hyperref}
\usepackage{nameref}
\usepackage{tabularx}
\usepackage{subcaption}
\usepackage{caption}
\usepackage{pifont} % For checkmarks and crosses
\usepackage{booktabs}
\usepackage{multirow}
\usepackage{array}
\usepackage{makecell} % necessario per multirowcell
\newcolumntype{C}[1]{>{\centering\arraybackslash}m{#1}}
\usepackage{tikz}
\usepackage{adjustbox}  % per resizebox
\usepackage{stmaryrd}
\usepackage{enumitem}  

\addto\extrasenglish{%
  
}

\usepackage{balance}
\usepackage[table]{xcolor}
\usepackage[T1]{fontenc}
\usepackage{float} % to use [H] option
\usepackage{acronym}
\usepackage{bm}
\usepackage{diagbox}

\newcommand{\rr}{\color{red}} %tolto
\newcommand{\bb}{\color{black}}
 
\newcommand{\bl}{\color{blue}} % aggiunto

\def\BibTeX{{\rm B\kern-.05em{\sc i\kern-.025em b}\kern-.08em
    T\kern-.1667em\lower.7ex\hbox{E}\kern-.125emX}}

\begin{document}

%\title{Multimodal Generative Prediction of Longitudinal NSCLC Tumor Progression}
\title{Longitudinal NSCLC Treatment Progression via Multimodal Generative Models}

\begin{comment}
\bl Dose-Aware Multimodal Generative Modeling of Longitudinal NSCLC Progression during Radiotherapy \bb

\bl Treatment-Aware Multimodal Generative Modeling of Longitudinal NSCLC Progression \bb

\rr Multimodal Generative Prediction of Longitudinal NSCLC Tumor Progression (previous)

\end{comment}

\author{

\IEEEauthorblockN{
Massimiliano Mantegna\IEEEauthorrefmark{1}\IEEEauthorrefmark{2},
Elena Mulero Ayll\'on\IEEEauthorrefmark{1},
Alice Natalina Caragliano\IEEEauthorrefmark{1},
Francesco Di Feola\IEEEauthorrefmark{3}
}

\IEEEauthorblockN{
Claudia Tacconi\IEEEauthorrefmark{4},
Michele Fiore\IEEEauthorrefmark{4}\IEEEauthorrefmark{5},
Edy Ippolito\IEEEauthorrefmark{4}\IEEEauthorrefmark{5},
Carlo Greco\IEEEauthorrefmark{4}\IEEEauthorrefmark{5},
Sara Ramella\IEEEauthorrefmark{4}\IEEEauthorrefmark{5}
}

\IEEEauthorblockN{
Philippe C. Cattin\IEEEauthorrefmark{6},
Paolo Soda\IEEEauthorrefmark{1}\IEEEauthorrefmark{3}\IEEEauthorrefmark{8}
Matteo Tortora\IEEEauthorrefmark{7}\IEEEauthorrefmark{8},
Valerio Guarrasi\IEEEauthorrefmark{1}\IEEEauthorrefmark{8}
}

\IEEEauthorblockA{\IEEEauthorrefmark{1}
Unit of Artificial Intelligence and Computer Systems,  
Department of Engineering, Università Campus Bio-Medico di Roma, Italy}

\IEEEauthorblockA{\IEEEauthorrefmark{2}
Multi-Specialist Clinical Institute for Orthopaedic Trauma Care (COT), Messina, Italy}

\IEEEauthorblockA{\IEEEauthorrefmark{3}
Department of Diagnostics and Intervention, Radiation Physics,  
Biomedical Engineering, Umeå University, Sweden}

\IEEEauthorblockA{\IEEEauthorrefmark{4}
Operative Research Unit of Radiation Oncology,  
Fondazione Policlinico Universitario Campus Bio-Medico, Rome, Italy}

\IEEEauthorblockA{\IEEEauthorrefmark{5}
Research Unit of Radiation Oncology,  
Department of Medicine and Surgery,  
Università Campus Bio-Medico di Roma, Italy}

\IEEEauthorblockA{\IEEEauthorrefmark{6}
Department of Biomedical Engineering,  
University of Basel, Allschwil, Switzerland}

\IEEEauthorblockA{\IEEEauthorrefmark{7}
Department of Naval, Electrical, Electronics and Telecommunications Engineering,  
University of Genoa, Italy}

\IEEEauthorblockA{\IEEEauthorrefmark{8}
These authors jointly supervised this work and share the last authorship.}

\thanks{Corresponding author: Valerio Guarrasi (valerio.guarrasi@unicampus.it)}
}
\maketitle

\begin{abstract}
Predicting tumor evolution during radiotherapy is a clinically critical challenge, particularly when longitudinal changes are driven by both anatomy and treatment. In this work, we introduce a \emph{Virtual Treatment} (VT) framework that formulates non-small cell lung cancer (NSCLC) progression as a \emph{dose-aware multimodal conditional image-to-image translation} problem.
Given a CT scan, baseline clinical variables, and a specified radiation dose
increment, VT aims to synthesize plausible follow-up CT images reflecting
treatment-induced anatomical changes.
We evaluate the proposed framework on a longitudinal dataset of
222 stage~III NSCLC patients, comprising 895 CT scans acquired during
radiotherapy under irregular clinical schedules. The generative process is
conditioned on delivered dose increments together with demographic and
tumor-related clinical variables. Representative GAN-based and diffusion-based models are benchmarked across 2D and 2.5D configurations.
Quantitative and qualitative results indicate that diffusion-based models benefit more consistently from multimodal, dose-aware conditioning and produce more stable and anatomically plausible tumor evolution trajectories than GAN-based baselines, supporting the potential of VT as a tool for in-silico treatment monitoring and adaptive radiotherapy research in NSCLC.
\end{abstract}


\section{Introduction}
The increasing integration of artificial intelligence (AI) into clinical
workflows is enabling a paradigm shift toward predictive and personalized
medicine, particularly in oncology, where decision-making relies on the joint
interpretation of imaging, clinical, and treatment-related data~\cite{schork2019artificial,tortora2023radiopathomics,caruso2024deep,paolo2026predicting,nibid2023deep, caragliano2025doctor,liu2021exploring}.
In this context, \emph{multimodal} learning has emerged as a key strategy for capturing patient-specific disease dynamics, especially in longitudinal scenarios where anatomical changes are driven not only by natural disease progression but also by therapeutic interventions~\cite{caragliano2025multimodal,guarrasi2025systematic,baumann2016radiation,CARUSO2025110843,cordelli2024machine}.
Radiotherapy represents a paradigmatic example of such a setting.
During treatment, patients routinely undergo baseline simulation CT for planning, followed by one or more follow-up CT scans acquired at irregular
intervals to monitor anatomical changes and treatment response.
However, longitudinal imaging in radiotherapy is inherently retrospective:
anatomical modifications are observed only after dose delivery, and acquisition schedules vary widely across patients due to toxicity, adaptive re-planning, and logistical constraints~\cite{bertholet2019real}.
These limitations hinder the ability to anticipate treatment-induced
changes and to explore alternative dose-response scenarios.

This challenge is particularly pronounced in non-small cell lung cancer
(NSCLC), which accounts for approximately 85\% of lung cancer cases and remains the leading cause of cancer-related mortality worldwide~\cite{kauczor2020esr}.
Longitudinal modeling in thoracic CT is further complicated by respiratory motion, variable lung inflation, and complex non-linear deformations involving both tumor and surrounding parenchyma~\cite{xiao2023deep}.
Despite these difficulties, the delivered radiation dose represents a measurable, controllable, and clinically meaningful driver of anatomical evolution, yet it is rarely incorporated explicitly as a conditioning variable in generative models.

Recent advances in generative AI have demonstrated the potential to synthesize
medical images with high fidelity, using Generative Adversarial Networks
(GANs)~\cite{TRONCHIN2026109234,mantegna2024benchmarking,guarrasi2025whole}.
While GAN-based image-to-image translation has been widely applied to medical imaging, most methods target static appearance changes or modality conversion and do not explicitly model longitudinal, treatment-driven anatomical evolution~\cite{loni2025review}.
More recently, Diffusion Models (DMs)~\cite{ho2020denoising} have shown promising results in modeling
spatiotemporal progression by conditioning generation on
temporal gaps and clinical variables~\cite{zhang2024development}.
However, existing approaches remain confined to neuroimaging, focus on natural aging rather than therapeutic effects, and do not exploit controllable treatment variables such as delivered dose~\cite{litrico2024tadm}.

In this work, we introduce a \emph{Virtual Treatment} (VT) framework that
formulates longitudinal NSCLC tumor evolution as a
\emph{dose-aware multimodal conditional image-to-image translation} problem.
Given a CT scan, patient-specific clinical variables,
and a radiation dose increment, VT aims to synthesize plausible follow-up CT scans that reflect treatment-induced anatomical changes.
The main contributions of this work are as follows:
\begin{itemize}
    \item We introduce the first
    \emph{Virtual Treatment} framework for radiotherapy that
    explicitly formulates NSCLC evolution prediction as a \emph{dose-aware multimodal conditional image-to-image
    translation} problem.
    \item We propose a tumor-focused loss that restricts the
    reconstruction objective to the Clinical Target Volume
    (CTV), using the baseline planning segmentation as an
    anatomical supervision that guides the models
    toward clinically relevant regions.
    \item We conduct a benchmarking study comparing GAN-based and diffusion-based generative models across 2D and 2.5D formulations, analyzing
    their volumetric tumor evolution and computational efficiency under dose-aware multimodal conditioning.
\end{itemize}

The remainder of the paper is organized as follows:~\autoref{sec:materials} describes the clinical dataset and acquisition protocol;~\autoref{sec:methods} presents the VT formulation, including training and inference strategies, and loss design;~\autoref{sec:experimental_setup} details the model architectures and implementation settings;~\autoref{sec:results} reports quantitative and qualitative results; finally,~\autoref{sec:conclusion} summarizes the findings and discusses future directions.

\section{Materials}
\label{sec:materials}

This study uses a private longitudinal dataset of 222 patients with stage~III NSCLC~\cite{ajcc2017} treated with radiotherapy at Fondazione Policlinico Universitario Campus Bio-Medico di Roma.
The cohort includes 895 thoracic CT scans, with one baseline simulation CT per patient and multiple follow-up CT scans acquired during treatment.
Radiotherapy followed standard protocols with daily
fractions of 1.8~Gy or 2.0~Gy, for a prescribed total dose of 60~Gy. 
Due to adaptive re-planning and protocol variations, the cumulative delivered dose reaches up to 68.4~Gy. Follow-up CT scans were acquired at irregular time
points, reflecting real-world clinical practice. 
On average,
each patient underwent $2.38 \pm 1.31$ follow-up scans.

For each acquisition, the cumulative delivered dose was recorded and used as the only longitudinal clinical variable. Baseline variables collected at treatment planning include age, sex, histological diagnosis, and clinical tumor and nodal staging (cT, cN). 
A CTV segmentation was available only for the baseline CT, as routinely performed in radiotherapy planning.

\subsection{Preprocessing}
\label{subsec:preprocessing}

All follow-up CT scans were registered to the corresponding baseline simulation CT using an affine and B-spline
transformation. 
CT intensities were converted to
Hounsfield Units, clipped to $[-900, 300]$~HU~\cite{joseph2025deep}, and normalized to
$[0,1]$. 
Volumes were resampled to a uniform voxel spacing of
$1\times1\times3~\mathrm{mm}^3$.
Spatial cropping was performed using a fixed in-plane bounding box derived from the largest baseline CTV in the cohort. Along the axial direction, we retained only slices containing the baseline CTV.

Clinical variables were standardized before training. 
Age and dose values were min--max normalized, sex was encoded as a binary variable, histological
diagnosis was one-hot encoded, and cT and cN were encoded as ordinal variables.

The resulting modalities provide consistent, anatomically focused inputs for training the generative models described in~\autoref{sec:methods}.

%%%
\begin{comment}
\begin{table}[t]
\centering
\renewcommand{\arraystretch}{1.4}
\setlength{\tabcolsep}{8pt}

\begin{tabular}{C{3cm} C{4cm}}
\hline
\textbf{Category} & \textbf{Variable} \\
\hline

% ------------------ DEMOGRAPHIC ------------------
\multirow{2}{*}{Demographic}
    & Age \\
    & Sex \\
\hline

% ------------------ TUMOR-RELATED ------------------
\multirow{3}{*}{Tumor-related}
    & Tumor diagnosis (7-histological subtypes) \\
    & Clinical tumor stage (T1--T4) \\
    & Clinical nodal stage (N0--N3) \\
\hline

% ------------------ LONGITUDINAL ------------------
Longitudinal
    & Radiotherapy dose \\
\hline

\end{tabular}

\caption{Clinical variables grouped by category.}
\label{table:clinical_variables}
\end{table}
\end{comment}

\begin{comment}
\begin{figure}[t] 
    \centering
    \includegraphics[width=0.9\linewidth]{curve_followup_ct.pdf}
    \caption{Distribution of follow-up CT scans across acquisition time points (CT\#), where CT\# denotes the sequential follow-up index after baseline}
    \label{fig:hist_ct}
\end{figure}

\begin{figure}[t]
    \centering
    \includegraphics[width=1.05\linewidth]{curve_followup_dose.pdf}
    \caption{Distribution of follow-up CTs across cumulative delivered dose intervals (Gy).}
    \label{fig:hist_dose}
\end{figure}
\end{comment}

\section{Methods} \label{sec:methods}
\begin{figure*}[t]
    \centering
    % --- Figura principale (sopra) ---
    \includegraphics[width=0.65\textwidth]{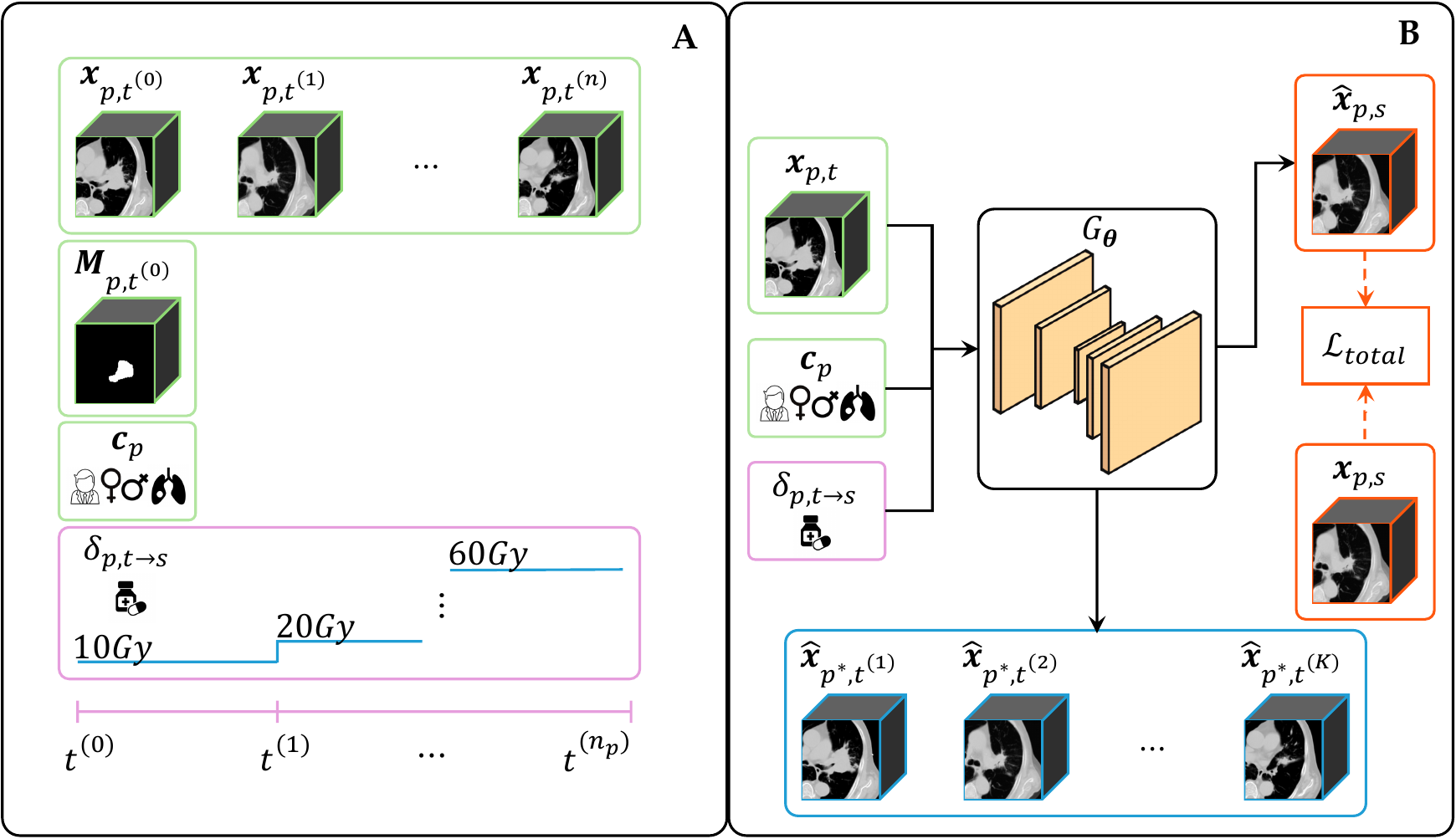}
    \caption{Overview of the proposed multimodal VT framework. \textbf{(A) Problem setting.} Multimodal conditioning for virtual treatment
    forecasting (\textit{green}: input CT and baseline clinical variables; \textit{pink}: dose increment).
    \textbf{(B) Generative formulation.} The generator $G_{\bm{\theta}}$ synthesizes
    follow-up CTs conditioned on input CT, clinical features, and dose increment,
    optimized with a reconstruction loss and a tumor-focused term within the CTV (\emph{orange}). Inference follows the \emph{direct dose-response} strategy
    (\emph{blue}) explained in~\autoref{sec:methods}.}
    \label{fig:vt_pipeline}
\end{figure*}
An overview of the proposed multimodal VT framework is shown in \autoref{fig:vt_pipeline}.
\textit{Panel~(A)} summarizes the problem setting, and \textit{Panel~(B)} outlines the training and inference procedures.

\subsection{Problem Setting}
We formalize the VT as a dose-aware multimodal conditional image-to-image translation task on longitudinal CT scans acquired during radiotherapy. Let $p\in\{1,\dots, P\}$ denote a patient observed at an irregular sequence of acquisition times:
\[
  \bm{\mathcal T}_p \;=\; \bigl\{t_{p}^{(0)} < t_{p}^{(1)} < \dots < t_{p}^{(n_p)}\bigr\},
\]
where $t_{p}^{(j)}$ is the acquisition time of the $j$-th CT scan, $t_{p}^{(0)}$ is the \emph{baseline} time point, and $n_p\!+\!1$ is the number of scans for patient $p$.
For simplicity, from this point onward we omit the patient index and write $t$ in place of $t_p$.

At $t^{(0)}$, a vector of baseline clinical features $\bm{c}_{p}\in\mathbb R^{d}$ is collected, where $d$ is the number of variables (age, sex, tumor
diagnosis, cT, cN). 
These features are reused as conditioning information for all acquisitions in $\bm{\mathcal T}_p$.  
At each $t\in\bm{\mathcal T}_p$, a CT volume $\bm{x}_{p,t}\in \bm{\mathcal X}\subset\mathbb R^{H\times W\times D}$ is available, where $H$ and $W$ denote the in-plane dimensions and $D$ denote the number of axial slices.

For any ordered pair $(t,s)$ with $t<s$ and $t,s \in \bm{\mathcal T}_p$, we define the \emph{dose increment} \[
  \delta_{p,t\to s} = d_{p,s}-d_{p,t},
\] 
which is the additional physical dose (Gy) delivered between two acquisitions.

\subsection{Virtual Treatment as a generative formulation}

Given a current observation $(\bm{x}_{p,t},\bm{c}_{p})$ and a dose increment $\delta_{p,t\to s}$, the goal of VT is to learn the conditional distribution
\[
  p_{\bm{\theta}}\!\,\bigl(\bm{x}_{p,s}\,\bigl|\,\bm{x}_{p,t},\bm{c}_{p},\delta_{p,t\to s}\bigr),
  \tag{1}\label{eq:generative_model}
\]
parameterized by $\bm{\theta}$, which denotes the learnable parameters of the 
chosen generative model.  
We model this distribution with an image-to-image generator $G_{\bm{\theta}}$ producing a prediction $\bm{\hat{x}}_{p,s}$ of the future scan $\bm{x}_{p,s}$. 
Formally:
\[
  \bm{\hat{x}}_{p,s} \;=\; G_{\bm{\theta}}\!\,\bigl(\bm{x}_{p,t};\bm{c}_{p},\delta_{p,t\to s}\bigr),
  \tag{2}\label{eq:generator}
\]

\textit{Training.}
From the training patients $\mathcal P_{\text{train}}$, we build a supervised dataset by enumerating \emph{all} ordered scan pairs. 
For each patient $p \in \mathcal P_{\mathrm{train}}$, and for every
future time point $s \in \bm{\mathcal T}_p$, we consider all earlier scans
$t \in \bm{\mathcal T}_p$ such that $t<s$, forming a training tuple
\[
  z = 
  \bigl(\bm{x}_{p,t},\, \bm{c}_p,\, \delta_{p,t\to s},\, \bm{x}_{p,s}\bigr).
\]
The resulting training set is therefore defined as
\[
  \mathcal D_{\text{train}} = 
  \left\{
     z:\,p\in\mathcal P_{\text{train}},\; t,s\in\bm{\mathcal T}_p,\,\; t\!<\!s
  \right\}.
  \tag{3}\label{eq:train}
\]
This construction ensures that, for each patient, the model is trained on the
complete set of admissible longitudinal transitions: by fixing a target scan
$s$ and collecting all compatible predecessors $t < s$, the dataset spans every
clinically observed evolution trajectory. As a consequence,
each mini-batch contains a heterogeneous mixture of temporal separations
$(s - t)$ and corresponding dose increments $\delta_{p,t\to s}$, naturally
exposing the model to a broad spectrum of progression patterns.

We define the \emph{multimodal conditioning vector}
\[
    \bm{h}_{\delta} = (\bm{c}_{p},\, \delta_{p,t\rightarrow s}),
\]
which concatenates baseline clinical variables and dose increment.
We use $\bm{h}_{\delta}$ in ~\autoref{sec:experimental_setup} to specify model conditioning.

\textit{Loss function.}
We employ a composite loss function $\mathcal{L}_{\mathrm{total}}$ that combines the standard objective of the generative model with a tumor-focused term:

\[
  \mathcal{L}_{\mathrm{total}} = \mathcal{L}_{\mathrm{main}} + \lambda \, \mathcal{L}_{\mathrm{tumor}},\tag{4}\label{eq:loss} 
\]
where $\mathcal{L}_{\mathrm{main}}$ denotes the training objective of the chosen architecture (GAN- or diffusion-based), $\mathcal{L}_{\mathrm{tumor}}$ encourages accurate reconstruction of the clinically most relevant region, and $\lambda$ is a hyperparameter controlling the 
relative weight of the tumor-focused regularization.

The main loss is averaged over all positions $i \in N$, where $N$ corresponds to the set of voxels in image space (GANs) or latent units in the noise prediction space (DMs):

\[
  \mathcal{L}_{\mathrm{main}}^{\mathrm{GANs}} 
    = \frac{1}{|N|} \sum_{i \in N} 
      \left| \bm{\hat{x}}_{p,s}^{(i)} - \bm{x}_{p,s}^{(i)} \right| \tag{5}\label{eq:loss_gan},
 \]     
\[ 
  \mathcal{L}_{\mathrm{main}}^{\mathrm{DMs}} 
    = \frac{1}{|N|} \sum_{i \in N} 
      \left| \bm{\hat{\epsilon}}_{p,s}^{(i)} - \bm{{\epsilon}}_{p,s}^{(i)} \right|,\tag{6}\label{eq:loss_dm}
\]
where $\bm{{\epsilon}}_{p,s}$ and $\bm{\hat{\epsilon}}_{p,s}$ denote the true and predicted noise in the diffusion process, respectively.

To guide learning toward anatomically meaningful changes, we introduce a tumor-focused $\ell_1$ loss restricted to the CTV of the target scan.
Let the tumor mask for patient $p$ at time $t$ be denoted by
\[
  \bm{M}_{p,t} \in \{0,1\}^{H \times W \times D},
\]
where $\bm{M}_{p,t}^{(i)} = 1$ if voxel $i$ belongs to the CTV and $0$ otherwise. The number of voxels within the tumor region is
\[
  |V_{\mathrm{CTV}}| = \sum_{i} \bm{M}_{p,t}^{(i)}.
\]
Given a target image $\bm{x}_{p,s}$ and its prediction $\bm{\hat{x}}_{p,s}$, the tumor loss is defined as
\[
  \mathcal{L}_{\mathrm{tumor}} = \frac{1}{|V_{\mathrm{CTV}}|}
  \sum_{i} \left| \bm{\hat{x}}_{p,s}^{(i)} - \bm{x}_{p,s}^{(i)} \right| \, \bm{M}_{p,t}^{(i)},\tag{7}\label{eq:loss_tumor}
\]
In this work, due to the limited availability of longitudinal tumor segmentations, we rely on a single tumor mask $\bm{M}_{p,t^{(0)}}$ obtained from the baseline planning CT and reused it across all subsequent time points of patient $p$. This choice is consistent with clinical practice, where the CTV is routinely delineated only on the simulation CT for treatment planning and not re-contoured on intermediate follow-up scans. 

\textit{Inference.}
For an unseen patient $p^\ast \in \mathcal P_{\text{test}}$, we adopt a
\emph{direct dose-response} strategy: starting from the baseline image
$\bm{x}_{p^\ast,t^{(0)}}$, we specify a sequence of \emph{dose increments}
\[
  \boldsymbol{\delta}_{p^\ast}
  = \{ \delta_{p^\ast,k} \}_{k=1}^K,
\]
each measured with respect to the baseline $t^{(0)}$. For each
$\delta_{p^\ast,k}$ value, the generator produces a dose-conditioned prediction
\[
    \bm{\hat{x}}_{p^\ast}^{(k)}
    =
    G_{\bm{\theta}}\bigl(
        \bm{x}_{p^\ast,t^{(0)}};\,
        \bm{c}_{p^\ast},\,
        \delta_{p^\ast,k}
    \bigr).
\]
Here, $K$ denotes the number of dose queries selected by the user, which does not necessarily correspond to the number of clinically acquired follow-up scans.
The model can be queried at arbitrary dose levels, including hypothetical values, thus generating a synthetic dose-indexed progression trajectory decoupled from the clinical acquisition schedule.

\section{Experimental Setup}
\label{sec:experimental_setup}

We evaluate the VT framework introduced in~\autoref{sec:methods} by comparing representative GAN-based and diffusion-based models within a unified dose-aware multimodal conditional image-to-image translation setting. 
All models take
as input a CT volume $\bm{x}_{p,t}$ and the conditioning vector $\bm{h}_{\delta}$ defined in~\autoref{sec:methods} and are trained to predict the future scan $\bm{\hat{x}}_{p,s}$ according to Eq.~\eqref{eq:generator}.
We benchmark two model families: (i) 2D GAN-based architectures (Pix2Pix and CycleGAN) and (ii) a 2.5D diffusion-based model (TADM), which represents state-of-the-art longitudinal forecasting under multimodal conditioning.

\subsection{Implementation and Training Details}
\label{subsec:comp_cost}

Experiments use a patient-level hold-out split with 80\% of patients for training, 5\% for validation, and 15\% for testing. Following Eq.~\eqref{eq:train}, we include all ordered scan pairs $(t,s)$ with $t<s$, yielding 78{,}502 slice-wise paired samples. 
We also include identity pairs ($t=s$) to stabilize near-zero dose predictions.

\subsection{2D GAN-Based Models}
\label{sec:gan_models}

We evaluate two representative 2D GAN-based image-to-image translation models, Pix2Pix and CycleGAN, as baselines for dose-conditioned CT synthesis. 
Both use a ResNet generator with an encoder-bottleneck-decoder structure and PatchGAN discriminators.
Pix2Pix is trained in a supervised setting with paired longitudinal CT slices. CycleGAN uses two generators and a cycle-consistency loss to support unpaired translation.  
In both cases, the conditioning vector $\bm{h}_{\delta}$ is encoded by a shared multi-layer perceptron and injected into the generator bottleneck as a
channel-wise modulation, enabling patient- and dose-specific control. 
Discriminators are kept unconditional to preserve an appearance-driven adversarial objective.
Both GANs are trained with an adversarial loss, an $\ell_1$ reconstruction term, and the proposed tumor-focused loss
(Eq.~\eqref{eq:loss_tumor}).

\subsection{2.5D Diffusion-Based Model}
\label{sec:tadm}
We adopt TADM~\cite{litrico2024tadm} as a diffusion-based baseline and adapt it to model treatment-driven, rather than age-driven, anatomical progression.

TADM operates in a 2.5D setting by processing triplets of consecutive axial CT slices. 
An encoder extracts anatomical context from the input triplet and injects it into a diffusion UNet for conditional denoising. 
Given an input scan $\bm{x}_{p,t}$, the model predicts a residual $\bm{\hat{r}}_{p,t\to s}$ and reconstructs the target scan as:
\[
\bm{\hat{x}}_{p,s} = \bm{x}_{p,t} + \bm{\hat{r}}_{p,t\to s}.
\]
Conditioning is implemented by embedding the dose increment $\delta_{p,t\to s}$ and the baseline clinical variables $\bm{c}_p$ through separate projection layers. 
The resulting embeddings modulate the UNet feature maps via additive feature-wise conditioning in each residual block, so that treatment-related and patient-specific information influence denoising throughout the network.
TADM is trained with the standard diffusion noise-prediction objective, combined with the tumor-focused loss (Eq.~\eqref{eq:loss_tumor}) applied after residual reconstruction.

\begin{comment}
\begin{figure}
 
    \centering
    \includegraphics[width=0.75\linewidth]{VT_main_figure_expsetup.pdf}
        %\caption{3D DM-based}
\caption{Architectural overview of the three generative configurations used in the multimodal VT framework. The pairwise conditioning vector is denoted as \(\bm{h}_{\delta}\), while the time-indexed conditioning vector is indicated as \(\bm{h}_{\tau}\). The panels show the architecture-specific pipelines for $G_{\bm{\theta}}$: 2D GAN-based, 2.5D DM-based, and fully 3D DM-based, which are described in detail in \hyperref[sec:models]{Section~\ref{sec:models}}
.}
\label{fig:vt_architectural overview}
\end{figure}  
\end{comment}

\section{Results} \label{sec:results}
In this section, we report the results of our multimodal VT framework for forecasting tumor evolution. All models were evaluated using tumor-specific volumetric error, computational efficiency indicators, and qualitative visual assessment.

All evaluation metrics were computed within a patient-specific \textit{Local} region, defined by the CTV bounding box. This region is clinically the most relevant for assessing treatment response, as it directly captures
dose-driven tumor evolution while excluding confounding anatomical variations in surrounding tissues, making it particularly appropriate for evaluating generative
forecasting performance.

\subsection{Tumor Segmentation Evaluation} \label{subsubsec:otsu_results}
To assess whether the models capture treatment-related tumor evolution, we
perform a tumor-focused volumetric analysis. Since manual CTV annotations are
available only at baseline, tumor extent at follow-up time points is estimated
using an Otsu-based thresholding strategy applied to both real and generated CT
scans.
For each follow-up scan, an Otsu-derived binary mask is extracted and tumor
volume is computed as the number of voxels within the thresholded region. The
relative volumetric discrepancy between predicted and real tumor volumes is
defined as:
\[
|\Delta V|
= 100 \cdot \frac{\lvert V_{\mathrm{pred}} - V_{\mathrm{real}} \rvert}
{V_{\mathrm{real}}},
\]
where $V_{\mathrm{pred}}$ denotes the tumor volume obtained from the Otsu mask of the generated scan, and $V_{\mathrm{real}}$ denotes the corresponding volume extracted from the real follow-up scan. This metric quantifies the relative deviation of the predicted tumor burden from the reference real volume.
\begin{figure}[t]
    \centering
    \includegraphics[width=\linewidth]{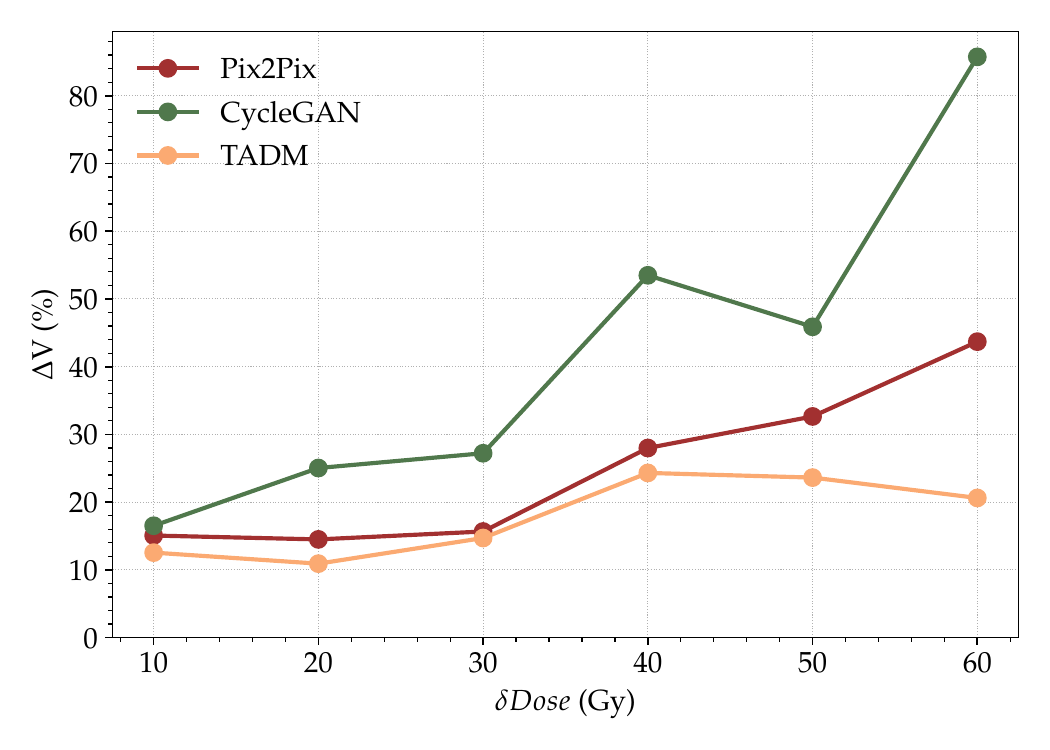}
    \caption{Percentage volumetric discrepancy $|\Delta V|$ between real and generated segmentations as a function of the dose increment ($\delta Dose$ (Gy)), aggregated into discrete dose bins (10--60~Gy). Each curve represents one generative model.}
    \label{fig:otsu_metric}
\end{figure}
    
The resulting percentage volumetric discrepancy trends, evaluated as a function of the cumulative dose increment, are reported in~\autoref{fig:otsu_metric}.

As shown by these trends, TADM produces more stable trajectories and moderate variation at higher dose levels.
In contrast, CycleGAN and Pix2Pix show the opposite behaviour, with a progressive overestimation of tumor shrinkage that becomes more pronounced for large dose increments.
The largest deviations are observed for CycleGAN, which exceeds 80\% error at 60 Gy.

Volumetric differences ($|\Delta V|$) within 25\% are considered clinically acceptable and are consistent with reported CTV variability in NSCLC radiotherapy (typically 7–25\%)~\cite{vu2020tumor}. 

Two conclusions emerge from this analysis:
(i) diffusion-based models preserve volumetric dynamics significantly better than GAN-based models and remain closer to real tumor evolution; (ii) all models show reduced accuracy for large dose differences, which confirms the difficulty of forecasting tumor progression over long temporal intervals.

\subsection{Computational Cost Analysis}
\begin{figure}[t]
    \centering
    \includegraphics[width=\linewidth]{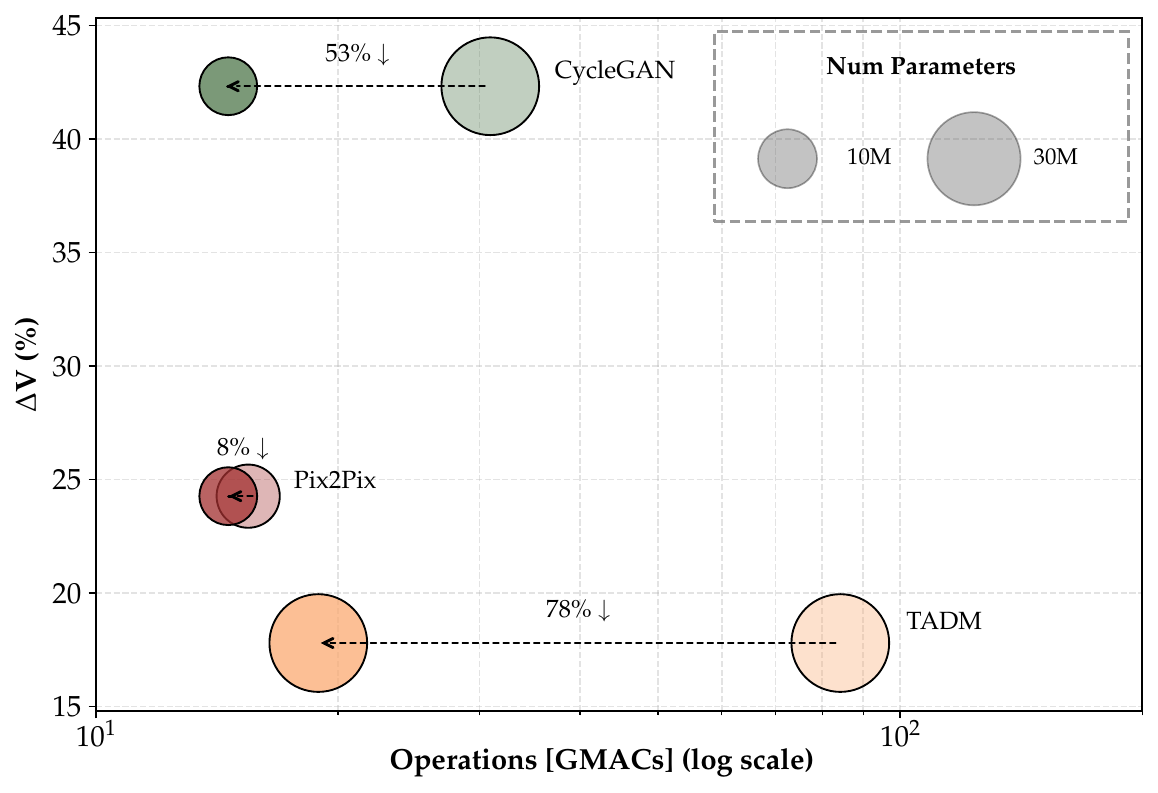}
\caption{Comparison of models in terms of $|\Delta V|$ error (y-axis) and computational cost measured in GMACs (x-axis, log scale) for the training (\emph{darker bubbles}) and inference (\emph{lighter bubbles}) phases. Bubble size is proportional to the number of model parameters (10M and
30M).}
\label{fig:bubbles}
\end{figure}

~\autoref{fig:bubbles} relates tumor volume error to computational cost,
measured in Giga Multiply-Add Operations per second (GMACs) and number of parameters. A clear accuracy-complexity
trade-off emerges: Pix2Pix and CycleGAN are lightweight but show higher volume
errors, whereas the diffusion-based TADM achieves substantially improved
accuracy at moderate computational cost.
Notably, the gap between training and inference cost differs markedly across
models. When transitioning from training to inference, the number of required
operations is reduced by approximately 53\% for CycleGAN, 8\% for Pix2Pix, and
78\% for TADM, highlighting a substantially more favorable deployment profile
for diffusion-based forecasting in longitudinal settings.
These results indicate that model evaluation for radiotherapy forecasting
should jointly consider predictive accuracy and computational efficiency,
particularly in longitudinal settings requiring large-scale virtual inference.

\subsection{Qualitative Results}
\label{subsec:qual_results}
\begin{figure}[t]
    \centering
    \includegraphics[width=0.5\textwidth]{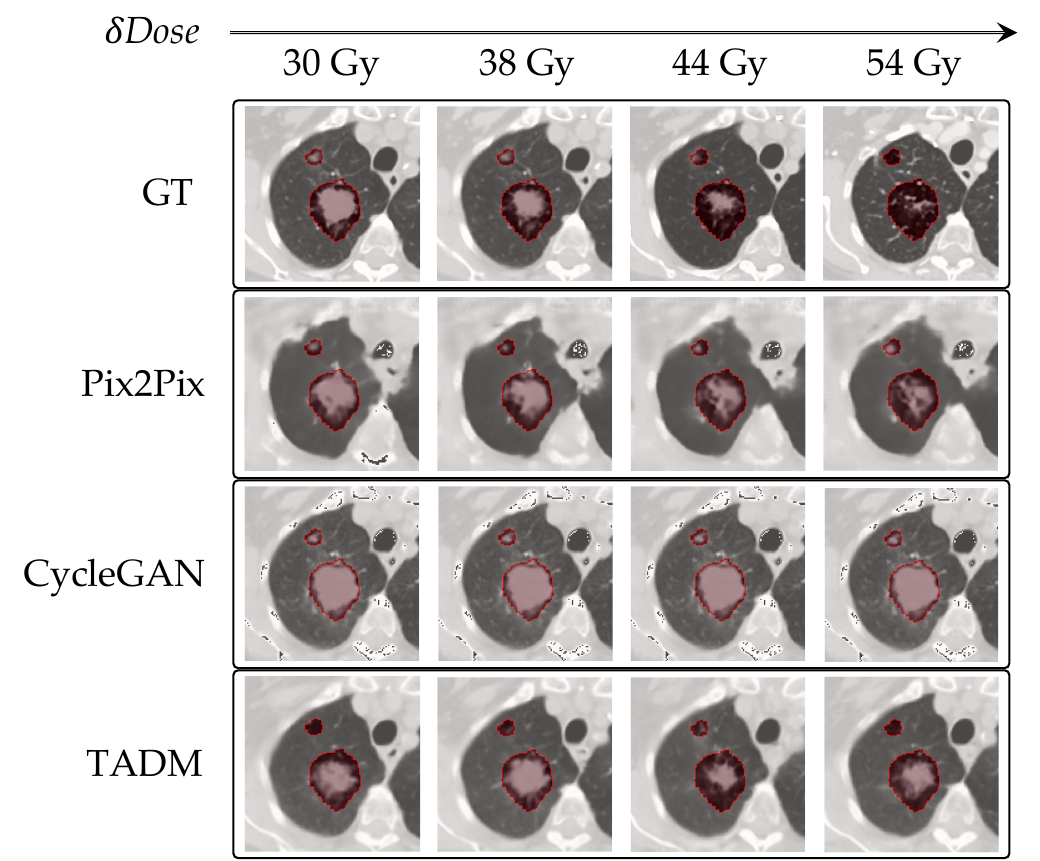}
    \caption{Qualitative comparison across increasing dose levels for a representative patient treated with a 2~Gy/day fractionation protocol. Columns correspond to increasing cumulative dose, while rows show the GT follow-up CT and the predictions generated by each model. The red contour indicates the CTV region used for \textit{Local} analysis.}
\label{fig:P}
\end{figure}

~\autoref{fig:P} shows representative qualitative results for a patient treated
with a 2~Gy/day fractionation scheme, illustrating model behavior across
increasing dose levels.
Predictions from each model are shown alongside the ground truth (GT), enabling a longitudinal visual assessment of tumor evolution
within the CTV region.
Consistent with quantitative findings, GAN-based models (Pix2Pix and CycleGAN)
show limited ability to capture dose-driven tumor regression. Pix2Pix tends to
preserve tumor appearance across dose levels, resulting in nearly static
trajectories and overestimation of residual tumor volume. CycleGAN further
introduces high-frequency artifacts and unstable local deformations, leading to
inconsistent volumetric trends.
In contrast, the diffusion-based TADM model captures a progressive,
dose-dependent reduction of tumor extent that closely follows the trend
observed in real follow-up CTs, indicating improved modeling of
treatment-induced anatomical change.
At high cumulative doses, diffusion-based predictions may overestimate tumor
regression, occasionally producing overly aggressive shrinkage. This behavior
reflects uncertainty in late-stage tumor morphology and highlights the
intrinsic difficulty of long-range longitudinal forecasting.

\section{Conclusions}
\label{sec:conclusion}
We presented a multimodal VT framework for forecasting
longitudinal CT evolution in NSCLC patients undergoing radiotherapy, comparing
2D GAN-based and 2.5D diffusion-based models under explicit dose-aware conditioning. 
To the best of our knowledge, this is the first study to systematically investigate treatment-driven anatomical forecasting in
thoracic CT with multimodal generative models.
Longitudinal tumor prediction proved inherently challenging due to irregular
acquisition schedules, non-uniform tumor regression, and the lack of
longitudinal annotations. Despite these limitations, diffusion-based models
demonstrated greater robustness and produced more stable volumetric predictions
across dose increments, whereas GAN-based models exhibited progressive error
accumulation and overestimation of tumor shrinkage at higher doses.
From a practical perspective, diffusion-based modeling achieved a more favorable
accuracy-efficiency trade-off, with TADM providing the best balance between
predictive stability and computational cost.
Future work will focus on integrating richer clinical descriptors, extending
the framework to fully 3D diffusion models, and incorporating physics-informed
priors and longitudinal segmentations~\cite{ayllon2025can,ayllon2025context} to further constrain
long-range tumor evolution. Overall, the proposed VT framework represents a promising step toward in-silico treatment monitoring and adaptive radiotherapy
support in NSCLC.

\section*{Acknowledgments}
Massimiliano Mantegna and Alice Natalina Caragliano are PhD students enrolled in the National PhD program in Artificial Intelligence, in the XXXVIII and XXXIX cycle, respectively, course on Health and life sciences, organized by Università Campus Bio-Medico di Roma. We also acknowledge partial financial support from: i) Università Campus Bio-Medico di Roma within the project ``AI-powered Digital Twin for next-generation lung cancEr cAre (IDEA)'', ii) JCSMK24-0094, iii) Cancerforskningsfonden Norrland project LP 25-2383.
Resources are provided by NAISS and SNIC at Alvis @ C3SE (grants no. 2022-06725 and 2018-05973).

\bibliographystyle{elsarticle-num}
\bibliography{references}

\end{document}